# First hardware demonstration of a continuous-state probabilistic computer with programmable energy-defined sampling

Fang Chao[1†], Hsuan Yao Tu[3†],Yihan He[1†*], Gong Xiao[1], Chin-Fu Nien[3*], and Gengchiau Liang[1,2*]

[1]Department of Electrical and Computer Engineering, National University of Singapore, 117583 Singapore

[2]Industry Academia Innovation School, National Yang-Ming Chiao Tung University, Hsinchu, Taiwan

[3]Department of Electronics and Electrical Engineering and Institute of Electrical and Computer Engineering, National Yang Ming Chiao Tung University, Hsinchu, Taiwan

[†]These authors contributed equally.

[*]Email: yihanhe@u.nus.edu; watchmannien@nycu.edu.tw; gcliang@nycu.edu.tw

***Probabilistic computing has proved useful for combinatorial optimization, but conventional implementations rely on binary stochastic units and sigmoid-based updates. Extending these systems to continuous variables requires binary encoding. Here we introduce a continuous probabilistic computer (CPC) whose continuous probabilistic bits sample bounded states from conditional distributions defined by local energies. Separating random-source generation from conditional sampling provides a route to using physical random sources with different native distributions. A field-programmable gate array (FPGA) prototype updates finite-precision states using digital randomness. The same framework supports product constraints, memory recovery and solution-space exploration. For the tested 7-bit factorization case, fixed-temperature FPGA CPC reduces mean core solving time approximately 33,000-fold relative to the evaluated quadratic unconstrained binary optimization software implementation. The FPGA recovers stored MNIST patterns without pretrained networks, achieving final success rates of 100%. Software CPC reaches 90% reference coverage of the feasible area in approximately 18% of the time required by restarted Adam.***

## Introduction ·

Energy-based computing provides a promising approach to solving combinatorial optimization problems by encoding objectives and constraints into an energy function. By programming interactions among computational units and local biases, the resulting energy landscape can be fast explored to identify low-energy configurations, with its ground states corresponding to optimal solutions for appropriately formulated problems. This principle underpins various physical computing paradigms, including quantum annealing and probabilistic computing (p-computing). Quantum annealing exploits quantum fluctuations and quantum dynamics to search for low-energy states of a problem Hamiltonian[3,4]. For example, a Rydberg-atom processor has found near-optimal independent sets on unit-disk graphs with up to 289 vertices[5]. In contrast, classical probabilistic computers (p-computers) are normally built upon the Ising model or Boltzmann machine[1,2] and employ networks of interacting binary-valued probabilistic bits (p-bits) to explore energy landscapes through stochastic state updates[6,7]. The binary nature of p-bits makes these p-computers particularly well suited to discrete optimization problems, while their intrinsic stochasticity and rapid sigmoidal switching dynamics enable fast exploration of the solution space through stochastic state updates. Recent advances in the co-design of p-bits based on emerging stochastic nanodevices and programmable coupling circuits[8,9], have demonstrated substantial acceleration across representative combinatorial optimization problems, including integer factorization, Max-Cut, the traveling salesman problem and Boolean satisfiability[10,11].

Despite these advances, many practical optimization problems involve inherently continuous variables, such as mixture coefficients, signal amplitudes, spatial coordinates and physical control parameters, whose optimal values are defined over continuous domains. Conventional binary p-bits are inherently restricted to two states and therefore cannot directly represent multistate or continuous-valued variables. The only possible approach is to combine multiple binary p-bits to approximate each variable with finite numerical precision, but it increases the number of computational units and the complexity of their interactions, potentially limiting the efficiency and scalability of optimization involving continuous variables. To overcome these limitations, recent efforts have sought to extend probabilistic computing beyond conventional binary representations. The probabilistic d-dimensional bit (p-dit) model generalizes binary p-bits to multidimensional states with real-valued components[12]. Gaussian probabilistic bits (g-bits) provide another approach by combining multiple binary p-bits to construct finite-precision variables with approximately Gaussian distributions, enabling experimental demonstrations of mixed binary-continuous optimization. However, These p-dit and g-bit demonstrations retain sigmoid-based discrete updates and do not directly sample native continuous device fluctuations. The demonstrated p-dit implementations have primarily focused on categorical and integer variables have yet to establish general continuous-variable optimization capabilities in hardware, while g-bits remain dependent on multiple binary stochastic units to represent each continuous-valued variable, introducing a trade-off between numerical precision and hardware resources. In parallel, thermodynamic computing exploits continuous stochastic dynamics to perform probabilistic computation, enabling continuous-variable sampling, matrix operations[14,15], with further developments extending their capabilities to nonlinear computation and learning[16,17]. However, unlike a programmable probabilistic bit, its target distributions are typically realized through engineered physical dynamics and interactions, making computational flexibility dependent on the underlying physical system. This motivates an alternative architecture in which continuous-valued stochastic variables are directly updated according to programmable, energy-dependent conditional distributions, with stochastic signal generation decoupled from probability distribution shaping.

Here we introduce a continuous probabilistic computer (CPC) built from continuous probabilistic bits (CP-bits), each representing a bounded continuous variable and sampling its state from a conditional distribution determined by its local energy (Fig. 1). An Ising-inspired coupling scheme defines the interactions between CP-bits, enabling different computational problems to be mapped onto a unified framework through programmable energy functions. A key feature of this architecture is the separation of stochastic signal generation from energy-dependent conditional sampling. Rather than requiring the physical random source to follow a specific target distribution, the CPC employs source-adapted sampling algorithms to transform native stochastic signals into samples from the desired conditional distributions. This design provides a flexible route towards incorporating different continuous stochastic devices, such as CMOS circuits exhibiting approximately Gaussian voltage fluctuations[18,19], and resistive random-access memory (RRAM) devices exhibiting log-normal resistance statistics without modifying the underlying energy formulation[20]. To our knowledge, the proposed framework is the first probabilistic computing framework designed to use native continuous device signals directly, adapting their distributions for energy-defined node updates.

To validate this framework, we develop a field-programmable gate array (FPGA)-based CPC prototype that integrates local energy evaluation, random-number generation, conditional sampling and sequential node updates. We demonstrate the versatility of the CPC framework across several computational tasks. In the decimal factorization task, the CPC efficiently identifies solutions to decimal product constraints, while also enabling stochastic exploration of feasible solution spaces. For associative memory, stored MNIST patterns can be recovered from noisy or incomplete inputs without pretrained networks. In continuous mixture decomposition, the CPC not only identifies compatible mixture coefficients but also explores multiple feasible solutions when the observations do not uniquely determine the underlying composition. These demonstrations show that the same CP-bit update framework can address both discrete-constraint problems and continuous-variable optimization, while enabling stochastic exploration of multiple feasible solutions. Together, these results establish CPC as a programmable probabilistic computing architecture for continuous-variable computation, with a pathway towards hardware implementations based on diverse stochastic devices.

## Results

### Continuous states and energy-defined local updates

CPC consists of interacting CP-bits. Each CP-bit has a state $x_i \in [l_i, u_i]$, whereas a conventional p-bit switches between two discrete states. Node biases and interactions define a common system energy[22,23,24].The energy with pairwise couplings between the first, second and third powers of node states is defined as:

$$E(\boldsymbol{x}) = E_0 + \sum_{p=1}^{3} (\boldsymbol{h}^p)^T \boldsymbol{x}^p + \frac{1}{2} \sum_{p,q=1}^{3} (\boldsymbol{x}^p)^T \boldsymbol{J}^{p,q} \boldsymbol{x}^q \tag{1}$$

where $E_0$ is a constant offset, $\mathbf{x}^p$ contains the $p$th powers of the node states, and $\mathbf{h}^p$ contains the corresponding bias coefficients. The term $J_{ij}^{p,q}$ weights ${x_i}^p {x_j}^q$, including mixed-power interactions such as $x_i {x_j}^2$ and ${x_i}^2 {x_j}^3$. The blocks satisfy $\boldsymbol{J}^{q,p} = (\boldsymbol{J}^{p,q})^T$ and contain no self-couplings (explain in Methods). The higher powers extend the linear biases and couplings used in Ising models[2,6]. At inverse temperature $\beta > 0$, the target Boltzmann distribution is

$$p_\beta(\boldsymbol{x}) = Z_\beta^{-1} exp[-\beta E(\boldsymbol{x})], \qquad \boldsymbol{x} \in \Omega, \qquad Z_\beta = \int_\Omega exp\,[-\beta E(\boldsymbol{y})]\, d\boldsymbol{y} \tag{2}$$

where $\Omega$ is the product of the node intervals, and $Z_\beta$ normalises the density. Lower-energy states have higher probability density, so task solutions can be encoded as energy minima. With the other node states $\mathbf{x}_{-i}$ fixed, the terms involving node $i$ form a local energy that is at most cubic in the candidate state $z$:

$$\begin{aligned} \varepsilon_i(z; \mathbf{x}_{-i}) &= a_i z^3 + b_i z^2 + c_i z \\ a_i &= h_i^3 + \sum_{q=1}^{3} \sum_{j \neq i} J_{ij}^{3,q} {x_j}^q, \\ b_i &= h_i^2 + \sum_{q=1}^{3} \sum_{j \neq i} J_{ij}^{2,q} {x_j}^q, \\ c_i &= h_i^1 + \sum_{q=1}^{3} \sum_{j \neq i} J_{ij}^{1,q} {x_j}^q. \end{aligned} \tag{3}$$

The remaining terms are constant with respect to $z$, giving $E(z, \mathbf{x}_{-i}) = \varepsilon_i(z; \mathbf{x}_{-i}) + \kappa_i(\mathbf{x}_{-i})$. Substituting this decomposition into equation (2) cancels $Z_\beta$ and $\kappa_i$, yielding the conditional probability density function (PDF):

$$\pi_i(z|\mathbf{x}_{-i}) = \frac{\exp[-\beta(a_i z^3 + b_i z^2 + c_i z)]}{\int_{l_i}^{u_i} \exp[-\beta(a_i v^3 + b_i v^2 + c_i v)] \mathrm{d}v}. \tag{4}$$

Constant, linear, convex quadratic and cubic local energies give uniform, truncated exponential, truncated Gaussian and cubic-energy non-Gaussian conditional PDFs, respectively (Fig. 1c). The sampler calculates $a_i$, $b_i$ and $c_i$ from the current states of the other nodes and then samples a new state for node $\boldsymbol{i}$ from the resulting conditional distribution in equation (4). Subsequent nodes use the updated state to calculate their local coefficients. A sweep updates each enabled node once. At fixed $\beta$, exact sequential conditional sampling constitutes Gibbs updating and preserves the target Boltzmann distribution[25,26].

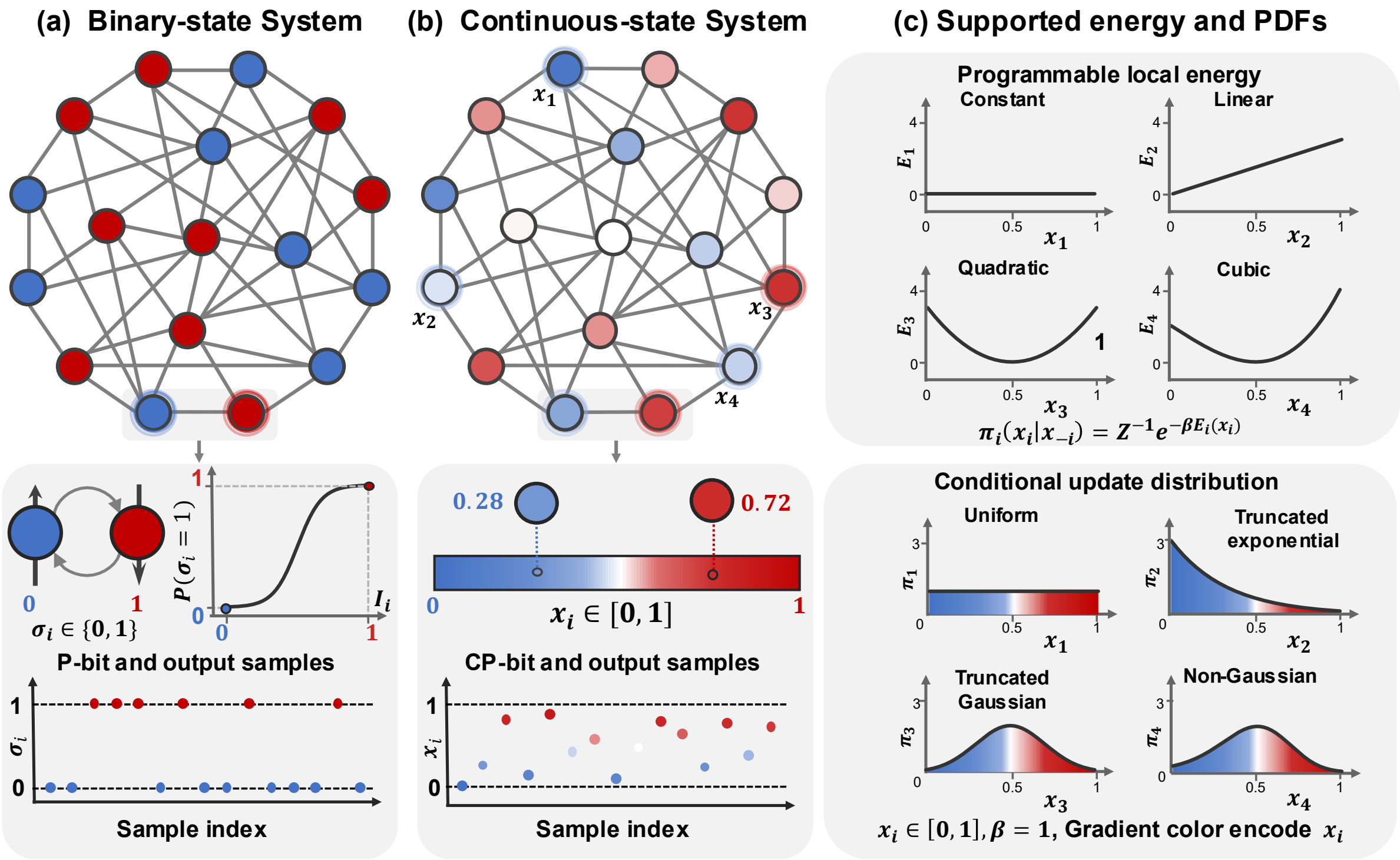

**Figure 1 | Continuous node states and energy-defined conditional distributions.** a, Binary-node network and p-bit output. b, Continuous-node network and CP-bit output on $[0,1]$. Colours indicate node states. c, Constant, linear, convex quadratic and cubic local energies (top) and the corresponding uniform, truncated exponential, truncated Gaussian and non-Gaussian conditional PDFs (bottom). Each node update draws a new state from the corresponding conditional PDF.

### FPGA implementation of continuous probabilistic computing

The FPGA prototype implements the local update rule in equations (3) and (4) on $[0, 1]$, with $a_i = 0$. For $b_i \geq 0$, the supported conditional family includes the uniform, truncated exponential and truncated Gaussian distributions described above. The application benchmarks primarily use truncated Gaussian conditionals, with uniform updates for degenerate product constraints.

Gaussian noise-source devices[27,28] could provide samples for this conditional family through suitable distribution transformations (Supplementary Note 7). Shifting and scaling Gaussian samples, followed by rejection outside the node interval, gives a truncated Gaussian distribution. For a characterised Gaussian source, its cumulative distribution function (CDF) maps samples to a uniform distribution. The inverse CDF of a truncated exponential then maps these uniform samples to that target distribution[29].

The FPGA prototype uses digitally generated pseudorandom numbers and a fixed-point rejection-sampling data path. At each sequential node update, the controller reads the bias and coupling parameters stored in block RAM (BRAM) and the latest neighboring states. The coefficient multiply–accumulate (MAC) module calculates $b_i$ and $c_i$ from equation (3), retaining only the first and second powers of neighbouring states. Fixed inputs contribute to these coefficients but remain unchanged. The inverse temperature $\beta$ enters the acceptance rule directly, allowing the same local-energy calculation to be used during fixed-temperature operation and annealing.

For the ideal quadratic conditional, the sampler first finds the minimum $m_i$ of the local energy on $[0, 1]$. For $b_i > 0$, the minimiser is $\frac{-c_i}{2b_i}$ clipped to this interval. For $b_i \leq 0$, the minimum is the smaller endpoint energy. This second branch also allows the mathematical sampling rule to handle negative quadratic coefficients, which are not used in the present benchmarks. Subtracting this minimum makes the energy difference non-negative and bounds the acceptance probability by one for non-negative inverse temperature. The architecture evaluates the negative of the temperature-scaled local energy and forms the equivalent difference between its maximum and the candidate value.

A linear-feedback shift register (LFSR) generates a uniform pseudorandom candidate $z$ on $[0, 1]$, and the random-number module supplies a uniform acceptance variate $u$ on $(0, 1)$. The sampler evaluates the temperature-scaled energy difference $\Delta E_i$ and uses a lookup table (LUT) to approximate $-\ln(u)$. In exact arithmetic, the acceptance rule is[29]

$$\begin{aligned} \Delta E_i &= \beta[\varepsilon_i(z; \mathbf{x}_{-i}) - m_i], \\ L_u &= -\ln(u), L_u \geq \Delta E_i. \end{aligned} \tag{5}$$

The derivation assumes mutually independent uniform variates that are independent of the pre-update system state. With exact arithmetic, the acceptance probability $e^{-\Delta E_i}$ gives the conditional density in equation (4) for $a_i = 0$ without evaluating its normalisation integral. After rejection, both random variates are regenerated while the neighbouring states, local coefficients and temperature remain fixed. Only the accepted state is written back for subsequent node updates.

The central schematic in Fig. 2 summarises coefficient calculation, temperature scaling, rejection sampling, state writeback and sequential scheduling. The controller repeats sweeps according to the temperature schedule and reads out states at specified times. Finite precision, random-number correlations and LUT errors can affect the implemented distribution. The left and right schematics in Fig. 2 show the inputs and outputs for the application benchmarks, which compare this FPGA implementation with software CPC based on direct or interpolated Gaussian inverse-CDF sampling. Methods and Supplementary Note 3 describe the software sampling procedures, and each benchmark caption specifies its timing definition.

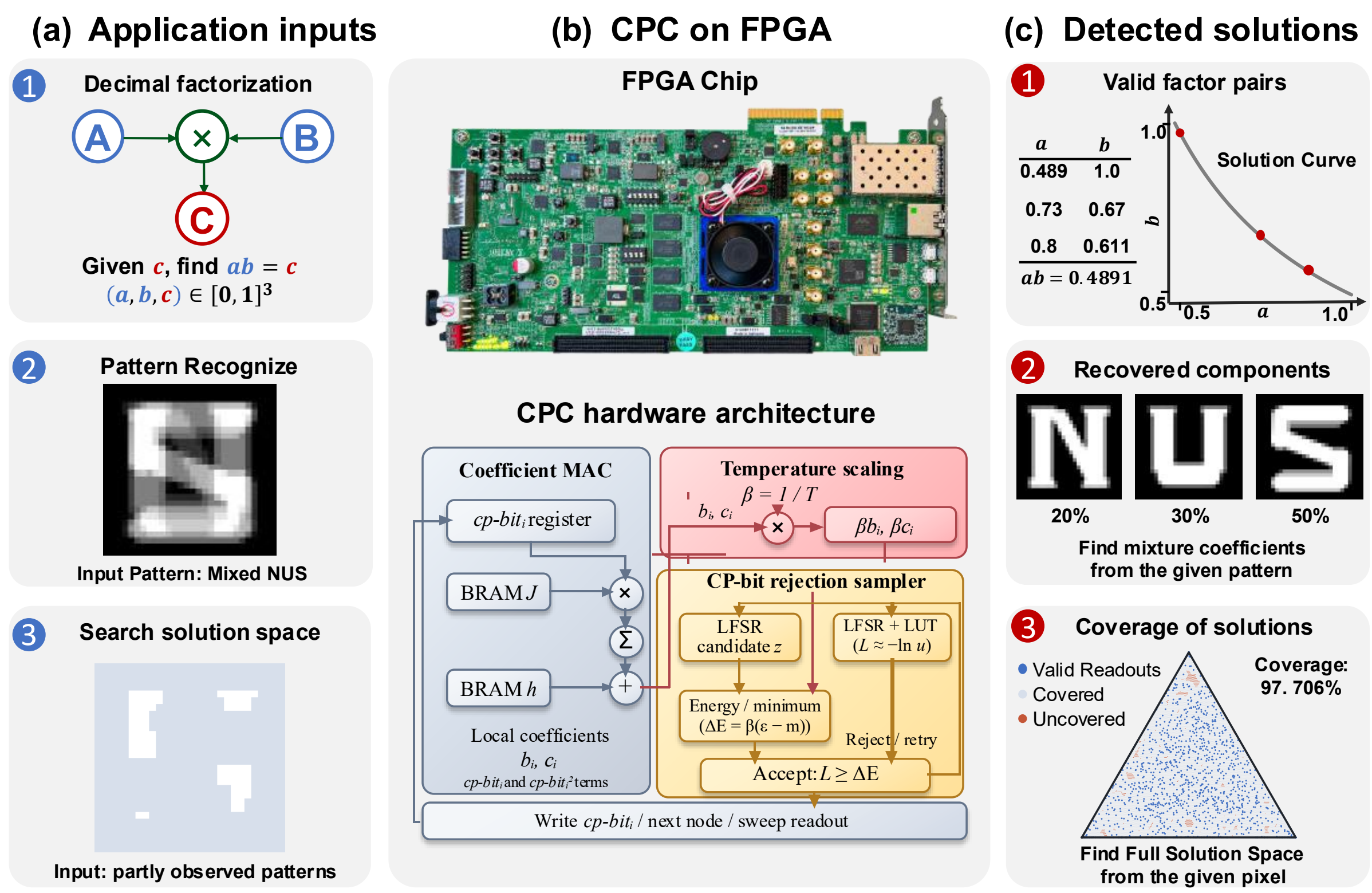

**Figure 2 | FPGA architecture and application workflow.** Left, inputs for continuous product constraints, pattern-mixture recovery and tasks with partial observations. Centre, FPGA board and CPC hardware architecture, including coefficient multiply–accumulate (MAC) operations, temperature scaling, LFSR candidate generation, LFSR–LUT acceptance-threshold generation, rejection sampling, state writeback and sweep readout. Right, product-constraint solutions, mixture coefficients and feasible solution sets. The application schematics illustrate the inputs and outputs of the common update process. Quantitative benchmark results are reported in Figs. 3–5.

**Continuous product constraints on the FPGA**

The continuous product constraint $ab = c$ tests whether the same update framework can find valid states and explore a continuous solution set. The states $a$ and $b$ lie in [0, 1], with $c$ fixed, and the energy $E = (ab - c)^2$ has minima along $ab = c$. Fixing either variable makes the energy at most quadratic in the other, allowing the conditional sampling rule described above. Supplementary Note 4 examines agreement the theoretical distribution derived from this energy with the empirical distributions generated by CPU and FPGA CPC.

All software methods were remeasured on the same CPU platform, with platform details given in Methods. An output is valid when $0.975c < ab < 1.025c$, and accuracy is defined as the fraction of valid outputs. Coverage is defined as the fraction of a discretised solution curve visited by valid outputs (Supplementary Note 4). Figure 3a uses 8 million recorded FPGA output pairs from 800 runs of 10,000 outputs each, with coverage calculated per run and then averaged. TTS and TTC90 denote the times to the first valid output and 90% coverage, respectively.

CPU comparisons assess the benefit of the continuous formulation. In the 7-bit exact-readout test (Fig. 3d), outputs scaled by S = 100 must round to the factors 67 and 73 with product 4891. CPC uses two continuous states, compared with 133 active binary nodes for QUBO. Fixed-temperature CPU CPC requires far fewer node updates and reduces mean core time approximately 3,000-fold. Node-update counts measure algorithmic work and do not imply equal arithmetic costs across methods. The advantage also depends on the task. At 10 bits, fixed-temperature CPU CPC reduces TTS and TTC90 by factors of 2.04 and 3.95 relative to restarted PGD (Fig. 3b,c), while CPU CPC annealing and PGD have similar mean times for the exact 7-bit readout. The specialised Solution-line method is faster for this exact readout because it constructs candidates analytically.

The FPGA implementation combines broad solution coverage with additional acceleration of CPC. In Fig. 3a, 10-bit FPGA annealing gives mean accuracy and solution-line coverage of 91.59% and 99.49%, respectively. Fixed-temperature FPGA CPC recovers the exact 7-bit factor pair in all tested runs, with a mean core time of 95.59 μs, 11.1 times faster than fixed-temperature CPU CPC. In the fixed-temperature 10-bit test, FPGA TTS and TTC90 are reduced by factors of 6.29 and 4.04 relative to fixed-temperature CPU CPC. The CPU and FPGA comparisons use the same problem and readout criteria, while each implementation retains its own sampler.

The current FPGA provides an initial demonstration of hardware feasibility, with further optimisation needed. FPGA annealing offers only a 1.22-fold time reduction relative to CPU CPC annealing for the exact 7-bit case, and its 10-bit coverage time is slightly longer than CPU annealing. The CPU sampler caches conditional parameters and uses incremental updates with direct Gaussian inverse-CDF sampling. The FPGA uses rejection sampling and recalculates local network inputs at each node update. Annealing and incremental input calculation remain targets for optimisation, although these measurements do not isolate their individual costs. The prototype uses digitally generated randomness, with physical random-source integration remaining future work.

For the tested 7-bit case, fixed-temperature CPC combines an approximately 3,000-fold CPU advantage over QUBO with a further 11.1-fold reduction from FPGA implementation, giving an overall core-time ratio of approximately 33,000. Using CPU CPC as the intermediate reference links the benefit of the continuous formulation on a common processor to the additional gain from the current FPGA implementation. These ratios apply to the tested implementations and the core timing boundaries defined in Methods.

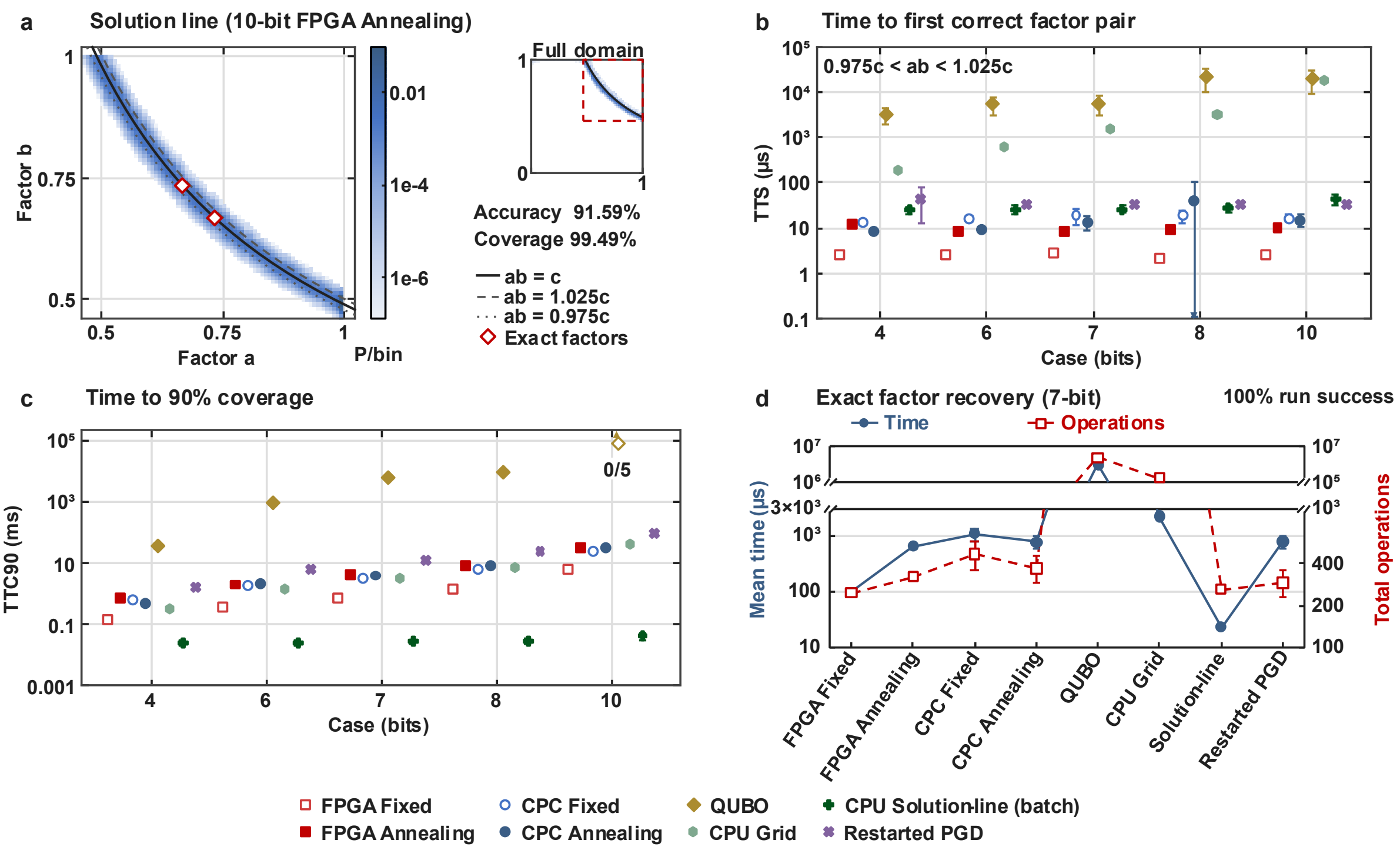

**Figure 3 | Continuous product constraints, solution-line coverage and exact integer readout.** a, Accumulated outputs from 10-bit FPGA annealing. Solid, dashed and dotted curves mark ab = c, ab = 1.025c and ab = 0.975c. Open diamonds mark the exact factor pair in both orders. The inset shows the full domain, and both distributions are normalised over all outputs. b,c, TTS and TTC90 across five case sizes. QUBO coverage is normalised by its encoding-reachable bins. The open QUBO marker and upward arrow in c indicate a time lower bound, with no coverage-threshold crossings in five runs (0/5). The downward arrow in b marks a confidence interval extending below the logarithmic axis. d, Mean core time (blue circles) and total operations (red squares) to the first exact 7-bit factor pair. Broken axes retain all data points and intervals. Points and error bars show means and 95% Student t confidence intervals. CPU timings use the 9800X3D measurements, and FPGA data are unchanged. Task criteria are defined in the text. Sample counts, operation counts and timing boundaries are given in Methods.

## Memory recovery and continuous mixture decomposition

Memory recovery extends the two-variable benchmark to larger networks while retaining the same energy-defined conditional updates (Fig. 4a). A dictionary of 200 MNIST patterns supports recall from a full noisy image and correction of an erroneous memory using 78 of 784 pixels. A 26-letter dictionary supports decomposition of a complete, noise-free mixture into five unknown components and their continuous weights. The dictionary and observations define the energy without network pretraining. Memory success requires the exact dictionary entry. Mixture success requires normalisable non-negative coefficients, a maximum image residual no greater than 0.005 and a coefficient half-L1 error no greater than 0.025.

On the common CPU platform, CPC reaches 100% final success in all three tasks and produces acceptable answers sooner than FCLS and Adam (Fig. 4b,d). Performance is measured by T50, the time at which the cumulative first-success fraction reaches 50%, with equal task weights and failed runs retained in the denominator. CPC T50 is 0.0857, 0.7878 and 0.1234 ms for recall, correction and mixture decomposition, respectively. These give time reductions by factors of 1.45–3.55 relative to FCLS and 3.05–11.37 relative to Adam under the tested implementations and readout rules. Nearest Memory is faster for correction but cannot recover continuous mixtures. FCLS and Adam also attain lower final image RMSE than CPC (Fig. 4c), distinguishing rapid acceptable recovery from final reconstruction precision given by CPC.

The FPGA retains high recovery success, but its recorded output times are longer than those of CPU CPC. Final success rates are 100% for all three tasks, with T50 values of 0.7667, 1.743 and 1.101 ms for the three tasks. Mixture recovery remains successful despite an increase in mean image RMSE from $2.53 \times 10^{-7}$ on CPU to $5.31 \times 10^{-4}$ on FPGA. Its recorded mixture T50 is below the Adam CPU value of 1.403 ms, although hardware task correspondence is inferred and equivalent end-to-end timing has not been established. These records therefore support successful hardware execution and a task-dependent latency comparison.

The longer FPGA output times relative to CPU CPC may be due to the implementation costs discussed above. CPU CPC uses compiled kernels, incrementally maintained matrix products and coefficient sums, and inverse-CDF sampling. The FPGA recalculates local inputs and uses rejection sampling, adding repeated arithmetic and sampling retries. Their separate contributions have not been profiled. The combined result is a common framework for memory identification and continuous composition, with short times to acceptable outputs on CPU and high success rates on hardware. Optimising input calculation and sampling could extend the hardware speed benefits demonstrated for product constraints to these larger networks.

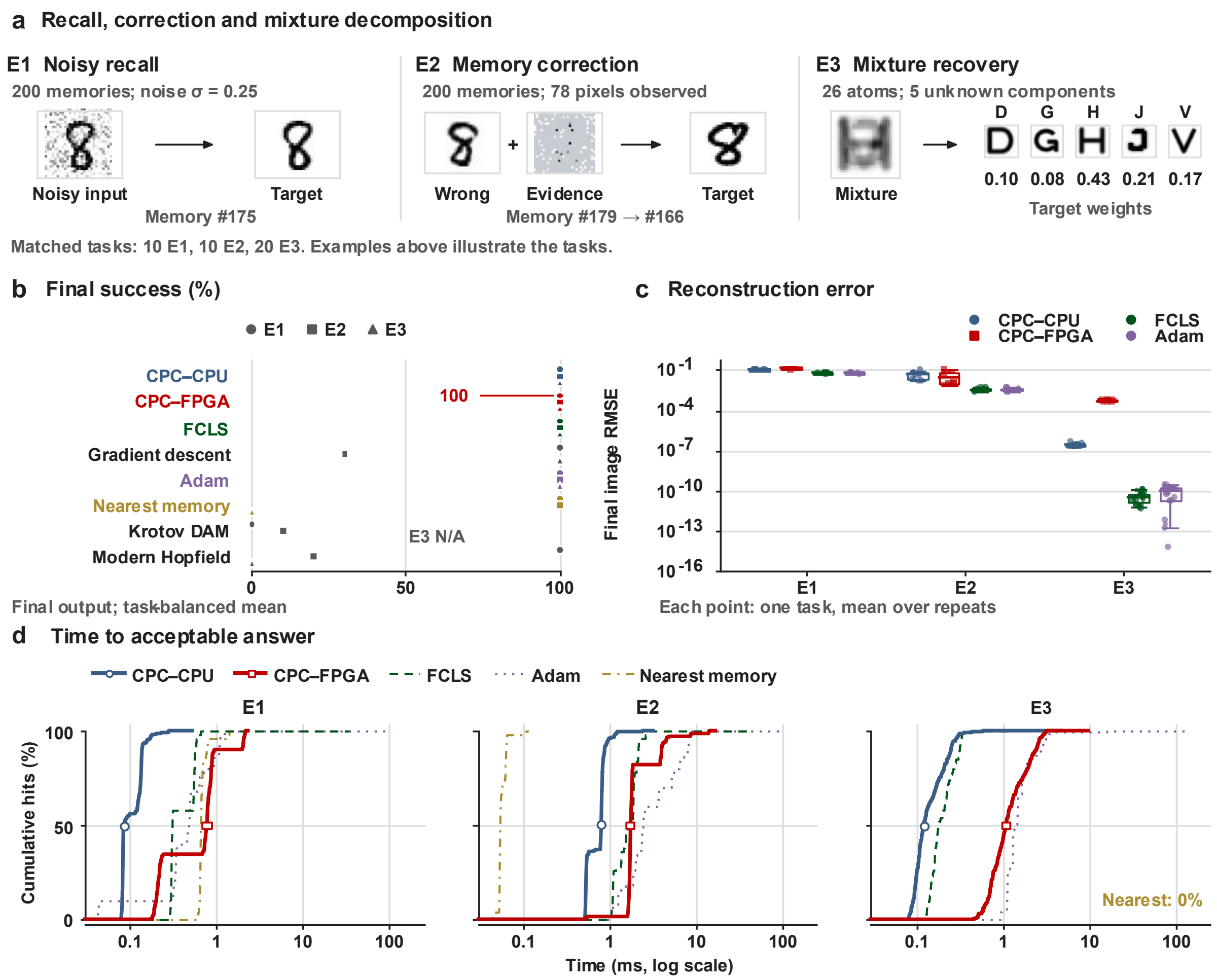

**Figure 4 | Memory recall, correction and continuous mixture decomposition.** a, Illustrative inputs and target outputs. Target identities and mixture weights are not provided to the solvers, and the correction example is outside the statistical cohort. b, Final success rates, with circles, squares and triangles denoting recall, correction and mixture decomposition. c, Final full-image RMSE. Points are task means over repeats, including failures. Boxes show the median and interquartile range, with 5th–95th percentile whiskers. d, Cumulative first-success fractions with equal task weights and failures retained. Open circles and squares mark CPC–CPU and CPC–FPGA T50. The cohort contains ten recall, ten correction and twenty mixture tasks. CPU data comprise 2,000 runs on the common platform, and FPGA data comprise 600 recorded runs. N/A denotes the absence of a native coefficient output. Software uses core runtime and FPGA uses recorded cycle timestamps. Hardware task correspondence is inferred, and equivalent end-to-end timing has not been established. Repeat counts and readout rules are given in Methods.

## Exploring observation-constrained solution spaces

Incomplete observations extend mixture recovery from finding one composition to exploring multiple compatible solutions. Fig. 5 uses the same 26-letter dictionary, with 108, 109 and 110 observed pixels restricting the feasible set to the N/U/S region (Space), the N–U segment (Line) and the N point (Pattern), respectively. CPU and FPGA use identical inputs and matching CPC energy weights. All 26 coefficients remain free to update. A normalised output is valid when its maximum observation residual is at most 0.005. Coverage counts equally weighted references within a full 26-dimensional half-L1 distance of 0.025 from a valid output. Space, Line and Pattern contain 20,100, 20,000 and one reference point, respectively, used only for scoring. TTS measures the first valid output and TTC90 measures 90% reference coverage.

CPU CPC explores these solution sets faster than restarted Adam with 5,000 steps per restart (Adam5000). Across ten seeds and 2,048 outputs per run, median TTC90 is 370.06 ms for Space and 126.25 ms for Line, giving 5.62-fold and 3.33-fold reductions relative to Adam5000 (Fig. 5c). CPC also has a shorter median TTS in all three tasks (Fig. 5e). Its median final coverage is slightly lower: 97.69% and 98.58%, compared with 98.52% and 100% for Adam5000. The comparison retains native output schedules, with two prescribed reads per CPC chain and Adam outputs released after each batch of 5,000-step restarts. Adam5000 also gives lower coefficient leakage, defined as normalised weight outside the target support averaged over all prescribed outputs, including invalid ones (Fig. 5d). Faster coverage therefore coexists with less selective allocation of coefficient weight.

The recorded FPGA batches reach 90.69% and 94.63% coverage for Space and Line, respectively, but do not establish a hardware speedup over CPU CPC. The chain-local counters is set at 10 ns per cycle gives serial-equivalent TTC90 estimates of 50.34 and 40.04 s, well above the CPU times. This reconstruction accumulates individual chain latencies, whereas the CPU implementation updates chains in vectorised batches and maintains local inputs incrementally. The FPGA input calculation and rejection-sampling costs described above may further contribute. The hardware results use one batch per task, with 2,560 Space outputs and 2,048 Line and Pattern outputs. These reconstructed times are not measured batch wall-clock times.

Independent chains provide a route to additional hardware acceleration. If all recorded chains start together with sufficient resources and no collection overhead, the same counters give estimated TTC90 values of 43.42 and 16.95 ms for Space and Line. These give time reductions by factors of 8.52 and 7.45 relative to CPU CPC, combining with the measured CPU advantage to give potential overall factors of 47.9 and 24.8 relative to Adam5000. This ideal parallel scenario preserves the coverage and leakage of the recorded outputs. The demonstrated advantage is rapid solution-set coverage on CPU and successful exploration on FPGA. Realising the projected additional gain requires a parallel implementation on FPGA and measured batch timing.

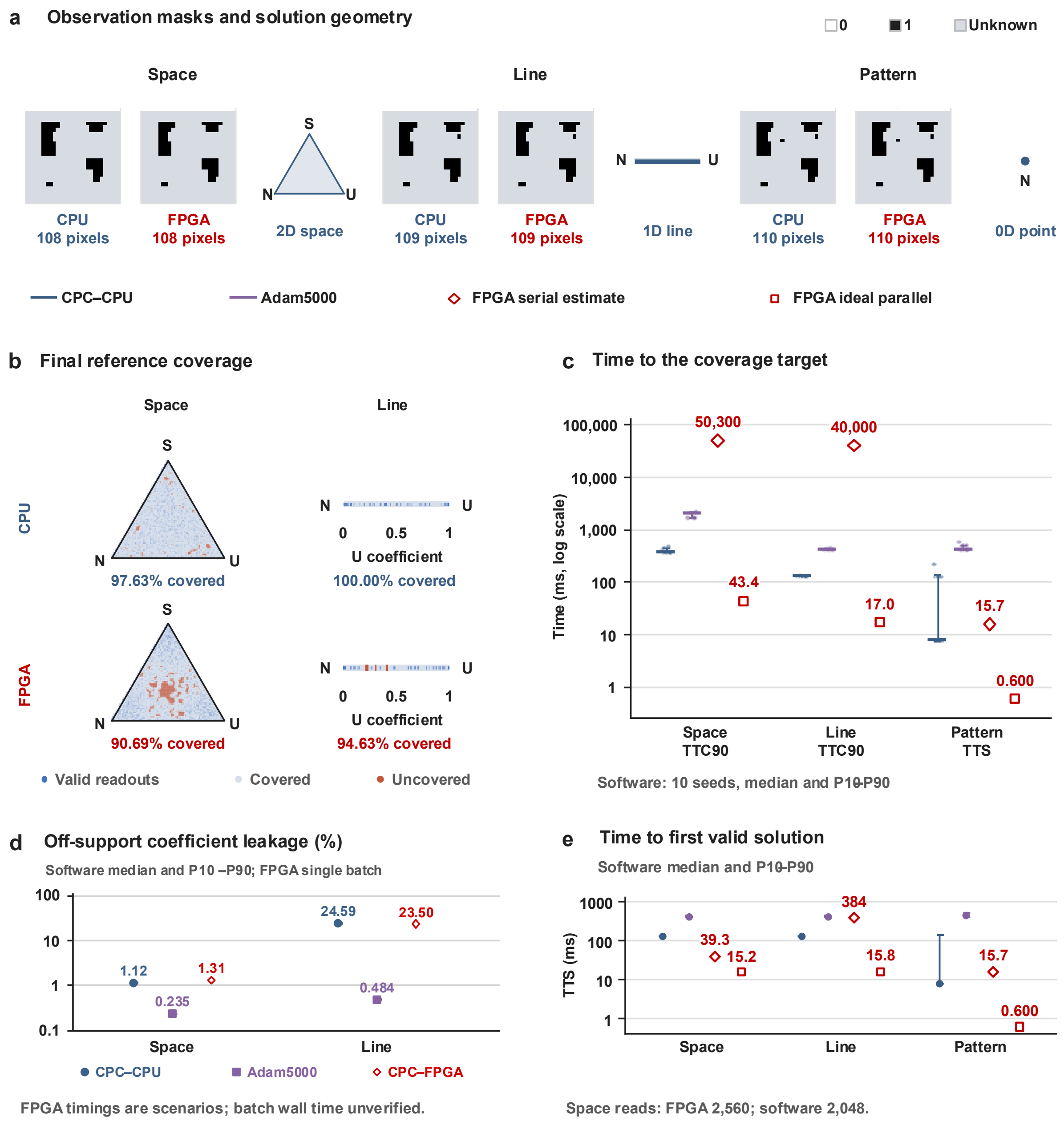

**Figure 5 | Coverage, coefficient leakage and timing under matched observation constraints.** a, Identical CPU and FPGA observation masks and the corresponding feasible sets. b, Coverage for CPU seed 1520001 and the complete FPGA batches. Pale and orange points denote covered and uncovered references, and dark points denote valid outputs. c, TTC90 for Space and Line, and TTS for Pattern. d, Normalised coefficient mass outside N/U/S for Space and N/U for Line, averaged over all prescribed outputs, including invalid ones. e, TTS, which coincides with TTC90 for Pattern in these data. Software statistics are medians and 10th–90th percentiles across ten seeds. FPGA coverage and leakage use one batch per task. Open red diamonds and squares show serial-equivalent and ideal parallel timing scenarios reconstructed at 10 ns per cycle. Neither is measured batch wall time. Each software run contains 2,048 outputs. FPGA uses 2,560 outputs for Space and 2,048 for Line and Pattern. Software timing includes preparation, initialisation, updates and output copying, and excludes offline scoring. Readout schedules are given in Methods.

## Conclusion

This work demonstrates a programmable CPC in which each CP-bit directly represents a bounded continuous variable, avoiding expansion into multiple binary optimisation nodes. This representation supports amplitudes, mixture weights and continuous constraints. The FPGA prototype stores each state at finite precision and updates its complete numerical value using digital randomness. Local energies define uniform, truncated exponential and truncated Gaussian conditional distributions, while cubic terms allow further non-Gaussian forms in the model. A major advantage of the architecture is its ability to accommodate different device random sources by separating random-source generation from conditional sampling. With source characterization and statistical adaptation, the framework can in principle use native continuous fluctuations, including approximately Gaussian signals, for energy-defined node updates. This provides a route beyond conventional sigmoid-based binary updates while retaining the same energy model and interaction rules. The same update framework supports product constraints, memory recall and correction, continuous mixture decomposition and solution-space exploration under partial observations. Stored patterns and observations define the memory-task energies directly, without a pretrained generative network. The FPGA achieves final success rates of 100% for recall, correction and mixture decomposition.

The benchmarks demonstrate rapid search for acceptable solutions and exploration of compatible solution sets. In the tested 7-bit exact-factorisation case, fixed-temperature CPU CPC reduces mean core solving time approximately 3,000-fold relative to the evaluated QUBO implementation on the same processor. FPGA execution gives a further 11.1-fold reduction to 95.59 μs, for an overall ratio of approximately 33,000. Across the three memory and mixture tasks, CPU CPC has a shorter T50 than FCLS and restarted Adam for reaching the prescribed success criteria. Those baselines achieve lower final reconstruction errors, and the current FPGA remains slower than CPU CPC in these tasks. When observations admit a solution region or line segment, CPC explores compatible continuous states while updating all 26 coefficients, with the target support and reference solutions used only for scoring. For the region and line segment, median software CPC times to 90% reference coverage are approximately 18% and 30% of the corresponding Adam5000 times respectively under the tested resolution, tolerances and readout rules. Optimising annealing, sampling, introducing incremental input calculation and integrating physical random sources are next steps for the FPGA prototype.

## Methods

### General energy model and local conditionals

The general CPC formulation uses a continuous system energy on the bounded state space $\Omega = \prod_{i=1}^{n}[l_i, u_i]$, with $l_i < u_i$. At fixed parameters $\boldsymbol{\theta}$, the energy can be written as

$$E(\mathbf{x}; \boldsymbol{\theta}) = \sum_{k=1}^{M} E_k\left(\mathbf{x}_{S_k}; \boldsymbol{\theta}_k\right). \quad (6)$$

In equation (6), $M$ is the number of energy terms, $S_k \subseteq \{1, \dots, n\}$ identifies the nodes involved in term $k$, and $\mathbf{x}_{S_k}$ is their state subvector. Each term $E_k$ describes a node bias or interaction with parameters $\boldsymbol{\theta}_k$. Terms may share variables and parameters[22,23,24]. This representation does not restrict the number of interacting nodes or the polynomial degree. Continuity of $E$ on the bounded closed domain $\Omega$ ensures $0 < Z_\beta < \infty$ in equation (2).

When a task fixes input nodes, their values remain unchanged throughout the run and enter the energy as given parameters. The target Boltzmann distribution is then defined over the remaining nodes for these input values. At fixed $\beta$, exact Gibbs updates preserve this distribution.

During an update of node $i$, the other states $\mathbf{x}_{-i}$ remain fixed. For a candidate value $z$, the local energy collects all terms involving node $i$, indexed by $F_i = \{k: i \in S_k\}$:

$$
\begin{aligned}
\varepsilon_i(z; \mathbf{x}_{-i}) &= \sum\nolimits_{k \in \mathcal{F}_i} E_k\big((z, \mathbf{x}_{-i})_{S_k}; \boldsymbol{\theta}_k\big), \\
E(z, \mathbf{x}_{-i}) &= \varepsilon_i(z; \mathbf{x}_{-i}) + \kappa_i(\mathbf{x}_{-i}).
\end{aligned}
\tag{7}
$$

The remaining terms form $\kappa_i$, which is independent of $z$. A node update therefore changes the local and system energies by the same amount. An interaction involving several nodes contributes to the local energy of each participating node. Substituting equation (7) into equation (2) gives

$$
\begin{aligned}
\pi_i(z|\mathbf{x}_{-i}) &= \frac{p_\beta(Z,\mathbf{X}_{-i})}{\int_{l_i}^{u_i} p_\beta(v,\mathbf{X}_{-i})\mathrm{d}v} \\
&= \frac{\exp[-\beta\varepsilon_i(z;\mathbf{x}_{-i})]}{\int_{l_i}^{u_i} \exp[-\beta\varepsilon_i(v;\mathbf{x}_{-i})]\mathrm{d}v}.
\end{aligned}
\tag{8}
$$

The global normalisation factor $Z_\beta$ and the term $\kappa_i$ cancel from this conditional density. To separate the cubic part from any additional dependence on $z$, write

$$
\varepsilon_i(z; \mathbf{x}_{-i}) = a_i z^3 + b_i z^2 + c_i z + d_i + r_i(z; \mathbf{x}_{-i}). \tag{9}
$$

The coefficients $a_i, b_i, c_i, d_i$ and residual $r_i(z; \mathbf{x}_{-i})$ depend on the fixed parameters and current states of the other nodes, and remain fixed during a conditional draw. The constant $d_i$ cancels from equation (8), but the residual must be retained unless $r_i \equiv 0$. When the local energy is exactly cubic or lower, equation (8) reduces to equation (4). This degree limit applies only to the updated variable. For example, ${x_i}^3 {x_j}^3$ has total degree six but is cubic in either variable when the other is fixed. The coupling matrices for the cubic model are detailed below, and sampling extensions are given in Supplementary Note 1.

**Conditional sampling and implementations**

The software reference for product constraints evaluates the Gaussian CDF and inverse CDF directly. Memory recovery evaluates the CDF with the error function and draws samples by inverse interpolation on a standardised grid. The mixture and observation-constrained solvers interpolate CDF and quantile tables. Each implementation uses the quadratic case of the conditional model in equation (4). Table ranges, endpoint handling, numerical fallbacks and readout rules are given in Supplementary Note 3.

**State representation and connections**

For $n$ nodes, equation (1) uses the vectors $\mathbf{x}^{(p)} = (x_1{}^p, \dots, x_n{}^p)^{\mathrm{T}}$ and $\mathbf{h}^{(p)} = \left(h_1^{(p)}, \dots, h_n^{(p)}\right)^{\mathrm{T}}$, with $p = 1,2,3$. The superscripts $(p, q)$ label the state powers at the two ends of a connection. The nine $n \times n$ coupling blocks form the $3n \times 3n$ matrix

$$\mathbf{J} = \begin{pmatrix} \mathbf{J}^{(1,1)} & \mathbf{J}^{(1,2)} & \mathbf{J}^{(1,3)} \\ \mathbf{J}^{(2,1)} & \mathbf{J}^{(2,2)} & \mathbf{J}^{(2,3)} \\ \mathbf{J}^{(3,1)} & \mathbf{J}^{(3,2)} & \mathbf{J}^{(3,3)} \end{pmatrix}. \tag{10}$$

The entry $J_{ij}^{(p,q)}$ is the coefficient of $x_i{}^p x_j{}^q$. For example, $J_{ij}^{(1,2)} x_i x_j{}^2$ couples a linear state to a squared state, while $J_{ij}^{(2,3)} x_i{}^2 x_j{}^3$ couples a squared state to a cubed state. Reverse blocks obey $\mathbf{J}^{(q,p)} = \left(\mathbf{J}^{(p,q)}\right)^{\mathrm{T}}$, making the full matrix symmetric. The blocks $\mathbf{J}^{(1,1)}$, $\mathbf{J}^{(2,2)}$ and $\mathbf{J}^{(3,3)}$ are individually symmetric, and the remaining blocks occur in transpose pairs.

Every block has zero diagonal entries, $J_{ii}^{(p,q)} = 0$, so a coupling always joins distinct nodes. Single-node terms are stored in the bias vectors. This separation keeps the energy at most cubic in each node. The factor 1/2 in equation (1) accounts for the two equivalent orientations of each interaction.

When node $i$ is updated, each neighbouring power $x_j{}^q$ is fixed, and $J_{ij}^{(p,q)} x_j{}^q$ contributes to the coefficient of $x_i{}^p$. The equal contributions from the reverse blocks combine with the factor 1/2 to give the coefficients in equation (3). All three local coefficients can therefore depend on the neighbouring states in the general cubic model.

The FPGA model retains only $p, q \in \{1, 2\}$, corresponding to $\mathbf{h}^{(3)} = \mathbf{0}$ and $\mathbf{J}^{(3,q)} = \mathbf{J}^{(p,3)} = \mathbf{0}$. The remaining coefficients follow from equation (3), with the sums restricted to the first and second powers, and the quadratic conditional density follows from equation (4). Fixed inputs contribute to the local coefficients through their connections. Details of fixed-input elimination are given in Supplementary Note 2.

### FPGA configuration and measurement

The FPGA designs were implemented on a KCU105 board with an XCKU040-FFVA1156-2-E device using Vivado 2025.2 and operated at 100 MHz. Each top-level design instantiates one CPC core and executes one chain at a time. For Fig. 3, states and local coefficients use signed 26-bit registers with 10 fractional bits. For Fig. 4, recall and correction use signed 24-bit state and coefficient storage with 10 fractional bits and a 40-bit sampling data path. Mixture decomposition also use 10 fractional bits, with 40-bit coefficient and a signed 63-bit sampling data path with 10 fractional bits. For Fig. 5, states use 10-bit fractional storage; coefficient storage is signed 48-bit with 10 fractional bits, and the scaled sampling datapath is signed 58-bit with 10 fractional bits. The rejection samplers use 32-bit acceptance-threshold LFSRs and 10-bit candidate LFSRs.

Hardware timing is derived from counters clocked at 100 MHz, with each count corresponding to 10 ns. In Fig. 3, a counter is cleared by the start pulse and runs continuously; its value is recorded with each valid factor-pair output. In Fig. 4, the counter starts with the task-start pulse and stops at output for recall, correction and mixture decomposition. Intermediate readouts also carry counter values. In Fig. 5, each chain counter starts when a valid chain-start request is accepted completing with final readouts. These intervals include initial-state loading, node updates and on-chip readout after the start request. FPGA configuration, board power-up, host control and transfer of captured data to the host are outside these counter intervals. Serial-equivalent and ideal parallel estimates for Fig. 5 are reconstructed from the per-chain counters as described in the main text.

### Product constraints and exact decoding

For Figs. 3a–c, outputs are scored using the validity criterion defined in the main text. Supplementary Note 4 gives the coverage discretisation and threshold rules. For Fig. 3d, Sa and Sb are rounded to integers with midpoint ties rounded upwards. Both the resulting factor pair and product must match the target, allowing either factor order.

Operation counts use active-node updates for CPC and QUBO, two variable updates per two-dimensional PGD step, and candidate evaluations for Grid and Solution-line. First-solution costs include failed attempts. Core timings exclude setup and offline scoring. The first exact readout is identified offline, so these times exclude online verification and return of the answer. Budgets are given in Supplementary Note 4.

Figure 3a pools 10,000 outputs from each of 800 FPGA annealing runs. For Fig. 3b,c, each case includes 100 runs per CPU CPC configuration, 800 per FPGA configuration, 20 each for QUBO and restarted PGD, and 10 each for Grid and Solution-line. The extended 10-bit QUBO coverage cohort contains five runs. Figure 3d uses a separate exact-readout cohort of 100 runs per CPU method and 800 per FPGA configuration, including the additional FPGA annealing runs. All runs in this cohort recover the exact factor pair. All Figure 3 CPU timings were remeasured with the frozen inputs, seeds, budgets and output rules, while the FPGA datasets were retained. The CPU archive contains 1,616 formal calls and 8,354,060 output pairs for Fig. 3b,c, plus 600 formal runs for Fig. 3d. The corresponding 690 and 30 warm-up calls are excluded.

### Energy mapping for memory recovery

The dictionary $\boldsymbol{P}$ and observations $\boldsymbol{y}$ are fixed, while the coefficients $\boldsymbol{w}$ vary within [0,1]. The observation set $\Omega_{\mathrm{obs}}$ selects the positions used in the reconstruction residual. This residual and a coefficient-sum penalty define the energy

$$E(\mathbf{y}, \mathbf{w}) = \alpha \left\| \mathbf{y}_{\Omega_{\mathrm{obs}}} - P_{\Omega_{\mathrm{obs}}} \mathbf{w} \right\|^2 + \lambda (\mathbf{1}^{\mathrm{T}} \mathbf{w} - 1)^2 \tag{11}$$

Here, $\alpha$ and $\lambda$ weight the residual and sum penalty, while $\beta$ denotes inverse temperature. Node updates apply neither simplex projection nor normalisation by the coefficient sum. Memory tasks scale the residual by observation count, whereas letter tasks use a fixed scale. Parameters and coupling expansions are given in Supplementary Notes 2, 5 and 6.

### Memory recovery, mixture decomposition and exploration

Figure 4 uses fixed dictionaries, observations and initial states for ten recall tasks, ten correction tasks and twenty mixture tasks. CPC on CPU and FPGA has 20, 20 and 10 repeats per task, respectively. CPU Adam and gradient descent have 5, 5 and 10 repeats, and each remaining CPU baseline has five. Hardware task correspondence is inferred from outputs, and the FPGA records do not establish independent random-seed replication. Memory identification clips the reconstructed image to [0,1] and selects the dictionary entry with the smallest squared image distance. At each temperature-stage boundary, software CPC releases the lowest-energy state among the initial state and completed stage endpoints. The mixture solver releases its current state after each sweep. First-success times use released outputs. For FPGA mixture readouts, the plotted time uses the first-packet counter, five cycles (50 ns) before the six-packet coefficient vector is complete. Noise levels, task indices and schedules are given in Supplementary Note 5.

For mixture and solution-space tasks, coefficients must be finite, non-negative and have a positive sum before offline normalisation and scoring against the thresholds in the main text. The residual threshold includes a numerical tolerance of $10^{-12}$. Scoring does not feed back into the chains. Figure 5 uses identical CPU and FPGA dictionaries, queries and observation masks, with matching ideal CPC energy weights for all three tasks. Each chain uses a fixed random node order. Reference sets are used only for offline scoring, and leakage statistics include invalid readouts. Construction and initialisation are detailed in Supplementary Note 6, and hardware timing is described in Supplementary Note 3.

For Fig. 5, CPC and Adam5000 each use ten seeds and release 2,048 outputs per run. CPC samples 1,024 chains with two reads per chain. Space and Line use a low–high–low temperature schedule and read at sweeps 32 and 96. Pattern uses a fixed low temperature and reads at sweeps 1 and 32. Adam5000 uses eight batches of 256 restarts and releases each batch after 5,000 steps. Software timing runs from solver entry to actual output release. The FPGA Space batch contains 1,280 chains, while Line and Pattern each contain 1,024, with two reads per chain.

**Baselines, timing and statistics**

The CPU results in Figs. 3–5 were remeasured on an AMD Ryzen 7 9800X3D under Windows 11, using Python 3.12.14, NumPy 2.3.5 and SciPy 1.18.1. Runs were executed serially with one numerical-library thread and recorded processor affinity. Warm-up runs were excluded. File box archive the respective code, frozen protocols, environments and per-run records. Each CPU comparison uses common inputs and scoring thresholds while retaining method-specific objectives and readout rules. Figure 4 CPC and mixture FCLS use compiled kernels. FCLS uses simplex-constrained least squares, and Adam uses different parameterisations in Figs. 4 and 5. Software timing covers core runtime for Figs. 3 and 4 and includes construction, initialisation, updates and output copying for Fig. 5. FPGA timing uses recorded timestamps. Figure 5 reconstructs serial-equivalent and ideal parallel scenarios at 10 ns per cycle. Timing boundaries are detailed in Supplementary Notes 3–6.

Independent runs or repeats within a task are the statistical units. Readouts from one chain are not independent experiments. Figure 3 reports means and 95% Student t confidence intervals, calculated from CPU timing blocks and FPGA runs. CPU groups generally use ten timing blocks, with five for the extended 10-bit QUBO coverage cohort. Figure 4 weights all tasks equally. Deterministic repeats describe timing variation and do not constitute independent random-seed validation. Figure 5 reports software medians and 10th–90th percentiles, with FPGA results from one batch. For Fig. 3 groups with runs that do not reach the coverage threshold, all runs are retained and times are truncated at a common observation window. The resulting mean is reported as a lower bound. Sample counts and interval calculations are given in Supplementary Table 3. No hypothesis tests were performed.

## References

1. Ackley, D. H., Hinton, G. E. & Sejnowski, T. J. A learning algorithm for Boltzmann machines. *Cogn. Sci.* **9**, 147–169 (1985). https://doi.org/10.1207/s15516709cog0901_7

2. Lucas, A. Ising formulations of many NP problems. *Front. Phys.* **2**, 5 (2014). https://doi.org/10.3389/fphy.2014.00005

3. Farhi, E. et al. A quantum adiabatic evolution algorithm applied to random instances of an NP-complete problem. *Science* **292**, 472–475 (2001). https://doi.org/10.1126/science.1057726

4. Albash, T. & Lidar, D. A. Adiabatic quantum computation. *Rev. Mod. Phys.* **90**, 015002 (2018). https://doi.org/10.1103/RevModPhys.90.015002

5. Ebadi, S. et al. Quantum optimization of maximum independent set using Rydberg atom arrays. *Science* **376**, 1209–1215 (2022). https://doi.org/10.1126/science.abo6587

6. Camsari, K. Y., Faria, R., Sutton, B. M. & Datta, S. Stochastic p-bits for invertible logic. *Phys. Rev. X* **7**, 031014 (2017). https://doi.org/10.1103/PhysRevX.7.031014

7. Singh, N. S. et al. CMOS plus stochastic nanomagnets enabling heterogeneous computers for probabilistic inference and learning. *Nat. Commun.* **15**, 2685 (2024). https://doi.org/10.1038/s41467-024-46645-6

8. Borders, W. A. et al. Integer factorization using stochastic magnetic tunnel junctions. *Nature* **573**, 390–393 (2019). https://doi.org/10.1038/s41586-019-1557-9

9. Duffee, C. et al. An integrated-circuit-based probabilistic computer that uses voltage-controlled magnetic tunnel junctions as its entropy source. *Nat. Electron.* **8**, 784–793 (2025). https://doi.org/10.1038/s41928-025-01439-6

10. Aadit, N. A. et al. Massively parallel probabilistic computing with sparse Ising machines. *Nat. Electron.* **5**, 460–468 (2022). https://doi.org/10.1038/s41928-022-00774-2

11. Chowdhury, S. et al. Pushing the boundary of quantum advantage in hard combinatorial optimization with probabilistic computers. *Nat. Commun.* **16**, 9193 (2025). https://doi.org/10.1038/s41467-025-64235-y

12. Duffee, C. et al. P-dits: probabilistic d-dimensional bits for extended-variable probabilistic computing. *Phys. Rev. Applied* **24**, 044077 (2025). https://doi.org/10.1103/4ngx-cmz7

13. Singh, N. S. et al. Beyond Ising: mixed continuous optimization with Gaussian probabilistic bits using stochastic MTJs. In 2024 IEEE International Electron Devices Meeting (IEDM) 1–4 (IEEE, 2024). https://doi.org/10.1109/IEDM50854.2024.10873478

14. Aifer, M. et al. Thermodynamic linear algebra. *npj Unconv. Comput.* **1**, 13 (2024). https://doi.org/10.1038/s44335-024-00014-0

15. Melanson, D. et al. Thermodynamic computing system for AI applications. *Nat. Commun.* **16**, 3757 (2025). https://doi.org/10.1038/s41467-025-59011-x

16. Whitelam, S. & Casert, C. Nonlinear thermodynamic computing out of equilibrium. *Nat. Commun.* **17**, 1189 (2026). https://doi.org/10.1038/s41467-025-67958-0

17. Lockwood, O. et al. A blueprint for equilibrium-based differentiable continuous-variable thermodynamic computing. Preprint at arXiv (2026). https://doi.org/10.48550/arXiv.2607.16183

18. Freitas, N., Delvenne, J.-C. & Esposito, M. Stochastic thermodynamics of nonlinear electronic circuits: a realistic framework for computing around kT. *Phys. Rev. X* **11**, 031064 (2021). https://doi.org/10.1103/PhysRevX.11.031064

19. Freitas, N. et al. Taming nonequilibrium thermal fluctuations in subthreshold CMOS circuits. *Phys. Rev. Applied* **25**, 034061 (2026). https://doi.org/10.1103/5l2z-3f1h

20. Karpov, V. G. & Niraula, D. Log-normal statistics in filamentary RRAM devices and related systems. *IEEE Electron Device Lett.* **38**, 1240–1243 (2017). https://doi.org/10.1109/LED.2017.2734961

21. Kingma, D. P. & Ba, J. Adam: a method for stochastic optimization. In International Conference on Learning Representations (ICLR) (2015). https://arxiv.org/abs/1412.6980

22. Bishop, C. M. *Pattern Recognition and Machine Learning* (Springer, 2006), pp. 386–387. https://doi.org/10.1007/978-0-387-45528-0

23. Wainwright, M. J. & Jordan, M. I. Graphical models, exponential families, and variational inference. *Found. Trends Mach. Learn.* 1, 1–305 (2008). https://doi.org/10.1561/2200000001

24. Kschischang, F. R., Frey, B. J. & Loeliger, H.-A. Factor graphs and the sum-product algorithm. *IEEE Trans. Inf. Theory* 47, 498–519 (2001). https://doi.org/10.1109/18.910572

25. Johnson, A. A., Jones, G. L. & Neath, R. C. Component-Wise Markov Chain Monte Carlo: Uniform and Geometric Ergodicity under Mixing and Composition. *Stat. Sci.* **28**, 360–375 (2013). https://doi.org/10.1214/13-STS423

26. Gelfand, A. E. & Smith, A. F. M. Sampling-based approaches to calculating marginal densities. *J. Am. Stat. Assoc.* **85**, 398–409 (1990). https://doi.org/10.1080/01621459.1990.10476213

27. Penzhorn, W. T. The design of a truly random monolithic noise generator. *Microelectron. J.* 15, 29–40 (1984). https://doi.org/10.1016/S0026-2692(84)80068-3

28. Hekmatshoar, B. & Shahidi, G. G. A low-power thin-film Si heterojunction FET noise amplifier for generation of true random numbers. *IEEE J. Electron Devices Soc.* 7, 388–392 (2019). https://doi.org/10.1109/JEDS.2019.2902183

29. Devroye, L. *Non-Uniform Random Variate Generation* (Springer, 1986). https://doi.org/10.1007/978-1-4613-8643-8